\documentclass[conference]{IEEEtran}
\IEEEoverridecommandlockouts
\usepackage[T1]{fontenc}
\usepackage{enumitem}
\usepackage{caption}
\usepackage{pgfplots}
\usepackage{array}
\usepackage{booktabs}
\usepackage{makecell}
\usepackage{amssymb}
\usepackage{cite}
\usepackage{amsmath,amssymb,amsfonts}
\usepackage{algorithmic}
\usepackage{graphicx}
\usepackage{textcomp}
\usepackage{xcolor}
\def\BibTeX{{\rm B\kern-.05em{\sc i\kern-.025em b}\kern-.08em
    T\kern-.1667em\lower.7ex\hbox{E}\kern-.125emX}}
    
\begin{document}

\title{\textbf{AMNS: Attention-Weighted Selective Mask and Noise Label Suppression for Text-to-Image Person Retrieval
\thanks{\textdagger: Corresponding author\\
\hspace*{2em}This work was supported by the Natural Science Foundation of China (No.62372082, U20A20184), Natural Science Foundation of Sichuan Province (No.2023NSFSC0485), the Fundamental Research Funds for the Central Universities (No.ZYGX2024Z017) and Shenzhen Natural Science Foundation (No.JCYJ20240813114206010).}
}
% {\footnotesize \textsuperscript{*}Note: Sub-titles are not captured for https://ieeexplore.ieee.org  and
}

\author{\IEEEauthorblockN{Runqing Zhang}
\IEEEauthorblockA{\textit{Shenzhen Institute for Advanced Study} \\
\textit{University of Electronic Science and Technology of China}\\
Shenzhen, China \\
202322280540@std.uestc.edu.cn}
\and
\IEEEauthorblockN{Xue Zhou\textsuperscript{\textdagger}}
\IEEEauthorblockA{\textit{Shenzhen Institute for Advanced Study} \\
\textit{University of Electronic Science and Technology of China}\\
Shenzhen, China\\
zhouxue@uestc.edu.cn}
}

\maketitle
\renewcommand{\footnoterule}{
    \hrule width 0.4\linewidth height 0.4pt
    \vspace{5pt}
}

\begin{abstract}
Most existing text-to-image person retrieval  methods usually assume that the training image-text pairs are perfectly aligned;  however, the noisy correspondence(NC) issue (i.e., incorrect or unreliable alignment)  exists due to poor image quality and labeling errors. Additionally, random masking augmentation may inadvertently discard critical semantic content, introducing noisy matches between images and text descriptions. To address the above two challenges, we propose a noise label suppression method to mitigate NC and an Attention-Weighted Selective Mask (AWM) strategy to resolve the issues caused by random masking. Specifically, the Bidirectional Similarity Distribution Matching (BSDM) loss enables the model to effectively learn from positive pairs while preventing it from over-relying on them, thereby mitigating the risk of overfitting to noisy labels. In conjunction with this, Weight Adjustment Focal (WAF) loss improves the model’s ability to handle hard samples. Furthermore, AWM processes raw images through an EMA version of the image encoder, selectively retaining tokens with strong semantic connections to the text, enabling better feature extraction. Extensive experiments demonstrate the effectiveness of our approach in addressing noise-related issues and improving retrieval performance.
\end{abstract}


\section{Introduction}
Text-to-image person retrieval utilizes textual descriptions to aid in the retrieval of person images \cite{b1,b31,b36}. Although many effective retrieval methods have been proposed in this field\cite{b2,b3,b4,b5,b34,b35,b37}, almost all of them implicitly assume that all input image-text training pairs are correctly aligned. However, in the real world, noise is prevalent due to factors such as environmental disturbances, imprecise linguistic descriptions, mislabeling, or missing information, etc., making it difficult for the idealized assumptions to hold. In addition, the random masking strategy\cite{b25,b32,b33} in the image enhancement process may incorrectly discard semantic content, leading to noisy matching between image patches and text descriptions, ultimately degrading retrieval performance. Therefore, effective noise suppression methods and image masking strategy are crucial for improving the model retrieval performance.

Among these challenges, the issue of noisy image-text pairs is particularly prominent. Mislabeled or low-quality images, treated as positive examples, mislead the model into learning incorrect associations, this phenomenon known as Noisy Correspondence (NC). In order to solve the NC problem, there are two main approaches: sample selection\cite{b7,b8,b26,b27} and adopting robust loss functions\cite{b9,b10,b29}. The former usually utilizes the memory effect of the deep neural networks\cite{b28} to progressively differentiate and filter noisy data, but this may require additional computational resources and carry the risk of mistakenly deleting valid data. Comparatively, robust loss functions provide a more direct and efficient solution to address NC.

Therefore, we focus on designing a noise-tolerant loss function to improve the model's robustness. Specifically, blindly pursuing a perfect match with all positive pairs can cause overfitting on noisy labels. To address this, we propose a novel cross-modal matching loss called Bidirectional Similarity Distribution Matching (BSDM) loss. By integrating the cosine similarity distribution of image-text pairs into both forward and inverse KL divergence, BSDM helps the model associate representations across modalities more flexibly, reducing sensitivity to noise. Additionally, we introduce a Weight Adjustment Focal (WAF) loss to further enhance the model's ability to handle hard samples, complementing BSDM in mitigating noisy labels' effects. In addition, to address the problem that random mask may destroy the semantic association between image patches and text description, we introduce an Attention-Weighted Selective Mask (AWM) strategy.  By computing attention weights for image regions, lower-weight regions are filtered out, allowing the model to focus on extracting more meaningful features. The AWM strategy aims to enhance feature extraction and retrieval performance, preserving image regions closely related to the text, while reducing noisy matches. The main contributions and innovations of this paper are as follows:
\begin{itemize}[leftmargin=*]
    \item We investigate two problems in text-to-image person retrieval tasks, namely the Noisy Correspondence (NC) and the noisy matching problem caused by random masking strategy. To our knowledge, this paper may be the first to explore the noisy matching problem caused by random masking strategy in pedestrian retrieval tasks.
    
    \item We propose the AMNS framework. Specifically, Attention-Weighted Selection Mask (AWM) strategy alleviates noise matching caused by random masks. Bidirectional Similarity Distribution Matching (BSDM) loss enables effective learning from positive pairs while avoiding over-relying, mitigating overfitting to noisy labels. Additionally, Weight Adjustment Focal (WAF) loss improves the model’s ability to handle difficult samples.
    
    \item Experiments show that AWM strategy is superior to random masking strategy. In addition, the noise suppression method proposed in this paper is applied to other text-to-image frameworks, and the performance is significantly improved. Among the AMNS methods, mAP index and mINP index have achieved superior performance.
\end{itemize}

% \subsection{Maintaining the Integrity of the Specifications}

% The IEEEtran class file is used to format your paper and style the text. All margins, 
% column widths, line spaces, and text fonts are prescribed; please do not 
% alter them. You may note peculiarities. For example, the head margin
% measures proportionately more than is customary. This measurement 
% and others are deliberate, using specifications that anticipate your paper 
% as one part of the entire proceedings, and not as an independent document. 
% Please do not revise any of the current designations.

\begin{figure}[t] % 使用 [t] 确保图像在页面顶部
\centering
\includegraphics[width=1\columnwidth]{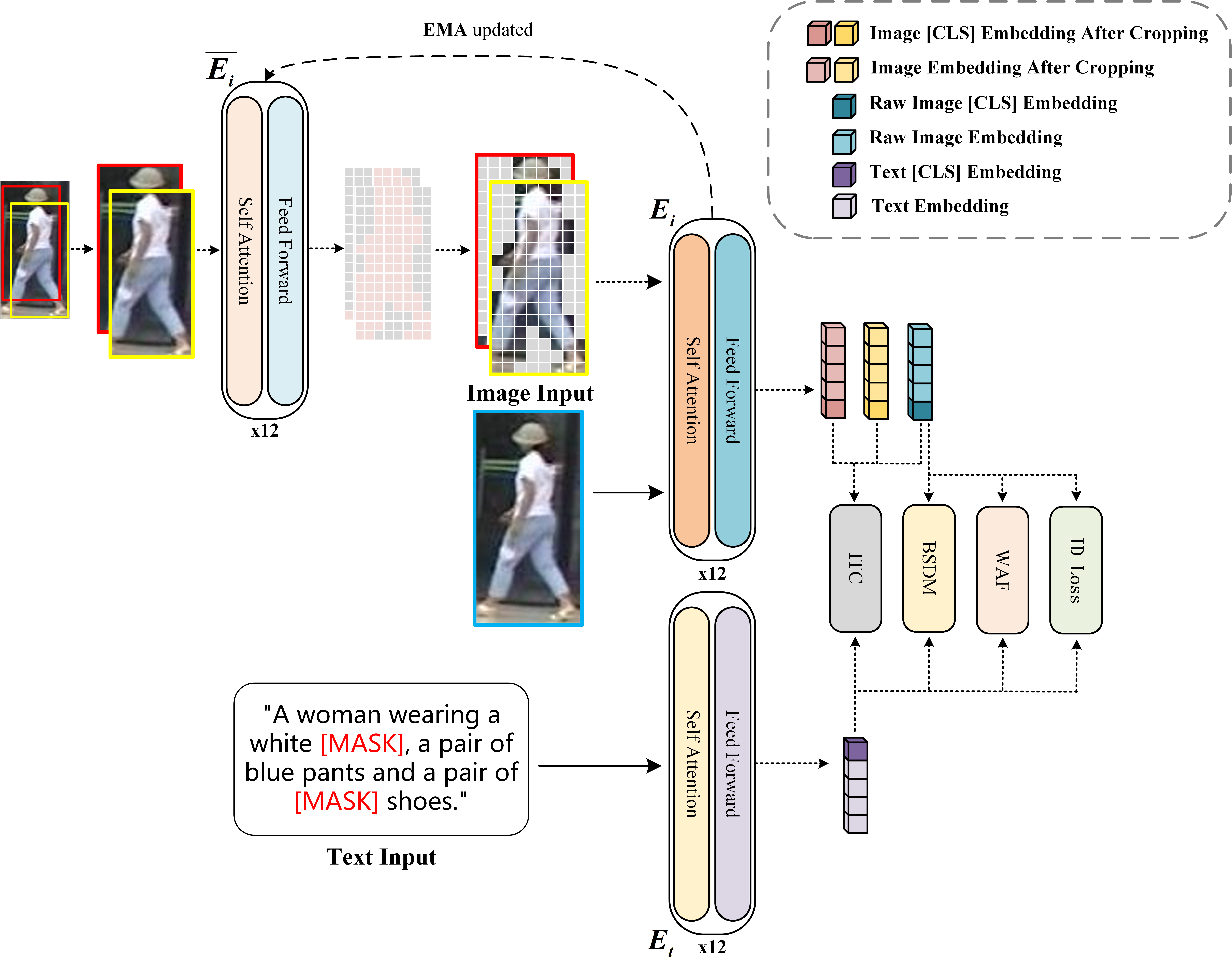} 
\caption{Overview of our proposed framework AMNS.}
\label{fig}
\end{figure}

\section{METHOD}
As shown in Fig. 1, our framework's overall structure includes image encoder ${\mathbf{E_i}}$, text encoder ${\mathbf{E_t}}$, and EMA-based image encoder $\overline{\mathbf{E_i}}$. The image is first cropped and processed by $\overline{\mathbf{E_i}}$ using AWM strategy to generate a mask that highlights regions relevant to the textual description. This masked image, combined with the original image, is then fed into ${\mathbf{E_i}}$ to extract more precise features. $\overline{\mathbf{E_i}}$ filters low-weight features via self-supervised learning, and with the proposed BSDM mitigates noisy label issues by aligning predictive and true distributions. Additionally, WAF enhances attention to hard samples by adjusting weight coefficients, ID loss aggregates the same identity features, further improving retrieval performance.

\begin{figure}[t] % 使用 [t] 确保图像在页面顶部
\centering
\includegraphics[width=1\columnwidth]{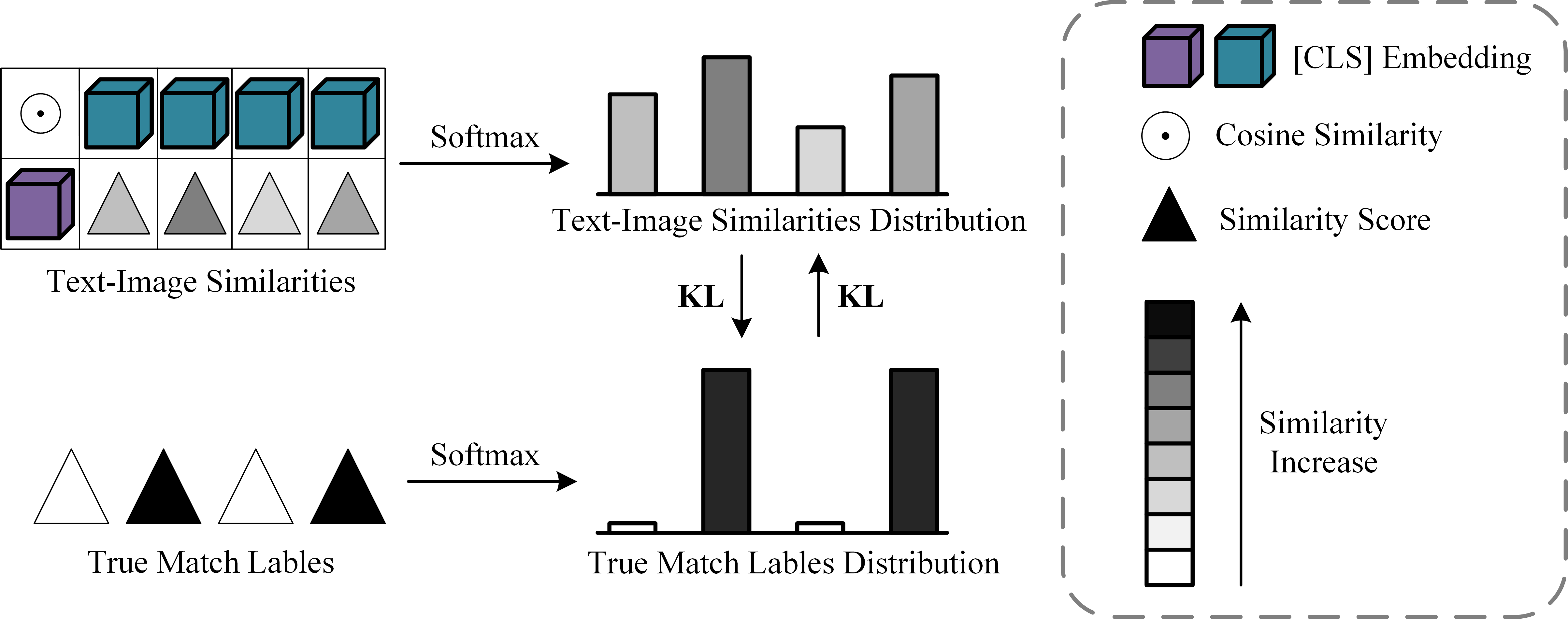} % 设置图像宽度为页面宽度的 70%
\caption{BSDM diagram. It demonstrates the core idea of implementing noisy label suppression. This approach constrains the true label distribution through the predicted distribution, preventing the model from becoming overconfident and mitigating the impact of noisy labels.}
\label{fig.2}
\end{figure}

% \subsection{Abbreviations and Acronyms}\label{AA}
% Define abbreviations and acronyms the first time they are used in the text, 
% even after they have been defined in the abstract. Abbreviations such as 
% IEEE, SI, MKS, CGS, ac, dc, and rms do not have to be defined. Do not use 
% abbreviations in the title or heads unless they are unavoidable.

\subsection{Attention-Weighted Selective Mask}
Since the $\mathit{[CLS]}$ token acts as a representative of global features, it is able to capture the overall semantics of the image and can correspond to the semantics contained in the textual modality. Using this property, we are able to evaluate the relevance of each image region to the overall semantics. Where the attention weight of the token located at position P is:
\begin{equation}
\mathit{A}_{\mathit{P}}=\frac{1}{HL}\sum_{l=1}^L\sum_{h=1}^H \text{Softmax}\left(\frac{\mathbf{f}_{lh}^q(\mathrm{CLS})\cdot\mathbf{f}_{lh}^k(P)}{\sqrt{d_k}}\right)
\end{equation}
where $\mathit{l}$ represents the layer index, and $\mathit{h}$ represents the attention head index. $\mathbf{f}_{\mathit{lh}}^q(\mathrm{CLS})$ refers to the query embedding of the $\mathit{[CLS]}$ token at Layer $\mathit{l}$ and Head $\mathit{h}$. $\mathbf{f}_{\mathit{lh}}^k(P)$ indicates the key embedding of Layer $\mathit{l}$ and Head $\mathit{h}$ for an image token at location $\mathit{P}$. The variable $d_{k}$ denotes the dimension of the key. In the overall mask strategy, the sizes of $\mathit{A}_{\mathit{P}}$ are sorted, and patches with lower weights are filtered out.

For a complete image, we introduce an Exponential Moving Average (EMA) version of the image encoder to generate the attention weights. At the same time, we use the Multiple Views strategy\cite{b12}, where we take two views and continue to introduce two self-supervised learning methods (SimCLR loss\cite{b13} and BYOL loss\cite{b14}) to assist in this mask task. Overall, the total loss of the AWM strategy is:
\begin{equation}
    L_{AW\!M}=L_{ITC}+L_{SimCLR}+L_{BYOL}
\end{equation}

\subsection{Noise Label Suppression Methods}
\label{BSDM}
SDM loss\cite{b4} effectively differentiates matching from non-matching pairs by aligning the cosine similarity distribution $\mathbf{p}$ of image-text pairs with the ideal label distribution $\mathbf{q}$. However, the SDM loss overly focuses on negative samples while neglecting the learning of positive samples.  Additionally, due to noisy correspondences in the dataset, forcibly fitting all positive sample pairs in the dataset leads to the model overfitting data containing noise.

Therefore, we propose Bidirectional Similarity Distribution Matching (BSDM), the core idea of which is shown in Fig.~\ref{fig.2}. BSDM allows the distribution p to also impose constraints on the distribution q. This bidirectional constraint provides the model with a certain degree of flexibility when fitting the data, enabling it to correctly fit positive sample pairs without blindly pursuing perfect alignment with all positive pairs, thereby helping to mitigate the impact of Noisy Correspondences (NC).
%折线图
\begin{figure}
\centering
\hspace*{-0.16\columnwidth} % 向左调整图的位置
\begin{tikzpicture}
\begin{axis}[
    title={},
    xlabel={Mask Ratio},
    ylabel={Rank-1},
    legend pos=north west,
    grid=major,
    grid style=dashed,
    width=0.9\columnwidth,
    height=5cm,
    ylabel near ticks,
    ymin=73, % Set the minimum value for y-axis
    ymax=75, % Set the maximum value for y-axis
    ytick distance=0.5, % Set the distance between ticks
    yticklabel style={/pgf/number format/fixed zerofill,/pgf/number format/precision=1},
    legend style={ % Adjust legend style here
        % at={(0,0)}, % Position the legend
        % anchor=north, % Anchor legend to the north
        legend columns=1, % Number of columns in the legend
        /tikz/every even column/.style={column sep=1ex}, % Space between columns
        /tikz/every even row/.style={row sep=1ex}, % Space between rows
        nodes={scale=0.8}, % Scale the nodes (text and markers)
    }
]
% Example data for m with solid blue circles
\addplot[
    color=blue, % Set the main color of the plot to blue
    mark=* , % Set the marker style to filled circle
    mark size=3pt, % Set the size of the marker
    mark options={fill=blue}, % Fill the marker with solid blue
    ]
    coordinates {
        (0.2, 73.75)
        (0.3, 74.06)
        (0.4, 74.14)
        (0.5, 74.25)
        (0.6, 74.19)
        (0.7, 74.20)
        (0.8, 74.03)
    };
\addlegendentry{AMNS$_{\mathrm{AWM}}$}

% Example data for $\tau$ with solid red triangles
\addplot[
    color=red, % Set the main color of the plot to red
    mark=triangle*, % Set the marker style to solid triangle
    mark size=3pt, % Set the size of the marker
    mark options={fill=red}, % Fill the marker with solid red
    ]
    coordinates {
        (0.2, 73.73)
        (0.3, 74.01)
        (0.4, 73.84)
        (0.5, 74.09)
        (0.6, 74.10)
        (0.7, 73.86)
        (0.8, 73.81)
    };
\addlegendentry{\hspace{0.8em}AMNS$_{\mathrm{Random}}$}

\end{axis}
\end{tikzpicture}
\caption{Rank-1 performance changes of the two strategies at different mask ratios.}
\label{Fig.3}
\end{figure}

First, given $N$ image-text pairs, $f_i^{v}$ and $f_j^{t}$ denote image global features and text global features. We construct a set of image-text representation pairs as $\{(f_{i}^{\nu}, f_{j}^{t}), y_{i,j}\}_{j=1}^{N}$, where $y_{i,j}=1$ represents matched pairs from the same identity, and $y_{i,j}=0$ represents unmatched pairs, $sim(\mathbf{u},\mathbf{v})=\mathbf{u}^{\mathsf{T}}\mathbf{v}/\|\mathbf{u}\|\|\mathbf{v}\|$ denotes the cosine similarity between u, v, and $\tau$ is the temperature hyperparameter. Then the probability of matching pairs can be calculated using the following softmax function.
\begin{equation}
    \label{eq.3}
    p_{i,j}=\frac{\exp{(sim(f_i^v,f_j^t)/\tau)}}{\sum_{k=1}^N\exp{(sim(f_i^v,f_k^t)/\tau)}}
\end{equation}

The image-to-text $\mathcal{L}_{i2t}$ loss in BSDM is computed by bidirectional KL scattering:
\begin{equation}
\label{eq.4}
\begin{aligned}\mathcal{L}_{i2t}&=KL(\mathbf{p_i}\parallel\mathbf{q_i})+KL(\mathbf{q_i}\parallel\mathbf{p_i})\\&=\frac1N\sum_{i=1}^N\sum_{j=1}^N\left(p_{i,j}\log\left(\frac{p_{i,j}}{q_{i,j}+\epsilon}\right)+q_{i,j}\log\left(\frac{q_{i,j}+\epsilon}{p_{i,j}}\right)\right)\end{aligned}
\end{equation}
where $\mathbf{\epsilon}$ is used to avoid very small numbers in numerical problems. $q_{i,j}=y_{i,j}/\sum_{k=1}^Ny_{i,k}$ is the true matching probability. $KL(\mathbf{q_i}\parallel\mathbf{p_i})$ is the forward KL scatter, and the goal is to make $\mathbf{p_i}$ as close to $\mathbf{q_i}$ as possible, this helps the model to learn the correct data representation. $KL(\mathbf{p_i}\parallel\mathbf{q_i})$ is the inverse KL scatter. With the bidirectional KL scattering, the model is made to be pushed not only to adapt to the real labels that may contain noise, but also to allow the real labels to adapt to the model's predictions, thus mitigating the effects of NC.

Correspondingly, the loss from text to image $\mathcal{L}_{t2i}$ can be exchanged for $f^{v}$ and $f^{t}$ in Eq.~(\ref{eq.3})~(\ref{eq.4}), BSDM loss is calculated by:
\begin{equation}
    \mathcal{L}_{BSDM}=\mathcal{L}_{i2t}+\mathcal{L}_{t2i}
\end{equation}
%三种损失在三个方法上的消融表格
\begin{table}[t]
\captionsetup{format=plain,justification=raggedright,singlelinecheck=false, font=small}
\caption{Comparison of the performance of the three losses SDM, TAL, and BSDM in the IRRA\cite{b4}, RDE\cite{b6}, and AMNS methods on the CUHK-PEDES dataset, where "Best" denotes the selection of the best checkpoints to be tested on the validation set, and the best results are shown in bold. The RDE method continues its performance comparison using two synthetic noise rates, i.e., 20\% and 50\% to simulate the performance under the real field where the image-text pairs are not well aligned.}
\centering
\setlength{\tabcolsep}{3pt} % 调整列间距
\fontsize{8}{9}\selectfont % 设置字体大小为8pt，行距为9pt
\resizebox{\columnwidth}{!}{ % 使用 \columnwidth 以填充整列
\begin{tabular}{c|cl|ccc|ccccc}
    \Xhline{1px}
    & \multicolumn{2}{c|}{} & \multicolumn{3}{c|}{Loss} & \multicolumn{5}{c}{CUHK-PEDES} \\
    Noise & \multicolumn{2}{c|}{Methods} & SDM & TAL & BSDM & Rank-1 & Rank-5 & Rank-10 & mAP & mINP \\
    \hline
    & \multicolumn{1}{m{1.5cm}}{IRRA$_{SDM}$} & Best & \checkmark & & & 73.38 & \textbf{89.93} & 93.71 & 66.13 & 50.24 \\
    0\% & \multicolumn{1}{m{1.5cm}}{IRRA$_{TAL}$} & Best & & \checkmark & & 73.20 & 89.02 & 93.34 & 66.05 & 50.49 \\
    & \multicolumn{1}{m{1.5cm}}{IRRA$_{BSDM}$} & Best & & & \checkmark & \textbf{73.51} & 89.38 & \textbf{93.75} & \textbf{66.27} & \textbf{50.71}\\
    \hline
    & \multicolumn{1}{m{1.5cm}}{RDE$_{SDM}$}  & Best & \checkmark & & & 75.26 & 89.91 & 94.02 & 67.37 & 51.40 \\
     0\% & \multicolumn{1}{m{1.5cm}}{RDE$_{TAL}$}& Best & & \checkmark & & 75.94 & 90.14 & 94.12 & 67.56 & 51.44 \\
    & \multicolumn{1}{m{1.5cm}}{RDE$_{BSDM}$}& Best & & & \checkmark & \textbf{76.06} & \textbf{90.38} & \textbf{94.36} & \textbf{68.04} & \textbf{51.99} \\
    \hline
    &\multicolumn{1}{m{1.5cm}}{RDE$_{SDM}$}& Best & \checkmark & & & 74.27 & 89.49 & 93.62 & 66.34 & 50.10 \\
    20\% &\multicolumn{1}{m{1.5cm}}{RDE$_{TAL}$}& Best & & \checkmark & & 74.46 & 89.42 & 93.63 & 66.13 & 49.66 \\
    &\multicolumn{1}{m{1.5cm}}{RDE$_{BSDM}$}& Best & & & \checkmark & \textbf{74.95} & \textbf{90.03} & \textbf{94.02} & \textbf{66.88} & \textbf{50.84} \\
    \hline
    &\multicolumn{1}{m{1.5cm}}{RDE$_{SDM}$}& Best & \checkmark & & & 69.33 & 86.99 & 91.68 & 61.99 & 45.34 \\
    50\% &\multicolumn{1}{m{1.5cm}}{RDE$_{TAL}$}& Best & & \checkmark & & \textbf{71.33} & \textbf{87.41} & 91.81 & 63.50 & 47.36 \\
    &\multicolumn{1}{m{1.5cm}}{RDE$_{BSDM}$}& Best & & & \checkmark & \textbf{71.33} & 87.30 & \textbf{92.20} & \textbf{63.75} & \textbf{47.51} \\
    \hline
    &\multicolumn{1}{m{1.5cm}}{AMNS$_{SDM}$}& Best & \checkmark & & & 73.73 & 89.18 & 93.55 & 67.35 & 52.88 \\
    0\% &\multicolumn{1}{m{1.5cm}}{AMNS$_{TAL}$}& Best & & \checkmark & & 73.39 & 88.97 & 93.13 & 67.17 & 52.82 \\
    &\multicolumn{1}{m{1.5cm}}{AMNS$_{BSDM}$}& Best & & & \checkmark & \textbf{74.25} & \textbf{89.54} & \textbf{93.83} & \textbf{67.67} & \textbf{53.16} \\
    \Xhline{1px}
\end{tabular}
}
\label{tab.1}
\end{table}

BSDM combines forward KL and inverse KL to balance each other, thereby mitigating the impact of NC, but it may weaken the model's ability to learn from hard samples (image-text similarity scores that are low, but are actually positive example samples, or similarity scores that are high, but are actually negative example samples). Therefore, we consider focal loss\cite{b11} and introduce proportional weight coefficients $\alpha$,$\beta$ in front of it to improve it, and by adjusting the coefficient factor $\gamma$ , we can adjust the model's attention to the misclassified samples. We refer to this loss as Weight Adjustment Focal Loss (WAF). The image-to-text WAF loss is as follows:
\begin{equation}
    \mathcal{L}_{W\!AF-i2t}=\begin{cases}
        -\alpha(1-p_{i,j})^\gamma \log p_{i,j}, & y_{i,j}=1 \\
        -\beta p_{i,j}^\gamma \log(1-p_{i,j}), & y_{i,j}=0
    \end{cases}
\end{equation}
Correspondingly, the total WAF loss is as follows:
\begin{equation}
    \mathcal{L}_{W\!AF}=\mathcal{L}_{W\!AF-i2t}+\mathcal{L}_{W\!AF-t2i}
\end{equation}

By combining BSDM and WAF, the model suppresses noisy labels and enhances learning of hard samples, achieving better performance in noisy data environments.

\textbf{Optimization}.We also employ the widely-used ID loss\cite{b30}. This loss aims to cluster feature representations of the same identity within the feature space, while feature representations of different identities are separated. This helps the model to better distinguish between different entities, thus improving the performance of model retrieval. The overall optimization objective of our training is defined as:
\begin{equation}
\mathcal{L}=\mathcal{L}_{AW\!M}+\mathcal{L}_{BSDM}+\mathcal{L}_{W\!AF}+\mathcal{L}_{id}
\end{equation}

% 第一个数据集比较表格
\begin{table*}[t]
\centering
\renewcommand{\arraystretch}{1.2} % 调整行间距
\caption{Performance comparison on three datasets. The best and second-best results are in bold and underline, respectively.}
\resizebox{\textwidth}{!}{

\begin{tabular}{c|c|ccccc|ccccc|ccccc}
\Xhline{1px}
 &   & \multicolumn{5}{c|}{CUHK-PEDES} & \multicolumn{5}{c|}{ICFG-PEDES} & \multicolumn{5}{c}{RSTPReid} \\
% \hline
 Methods & Ref. & Rank-1 & Rank-5 & Rank-10 & mAP & mINP & Rank-1 & Rank-5 & Rank-10 & mAP & mINP & Rank-1 & Rank-5 & Rank-10 & mAP & mINP \\
\hline
IVT\cite{b18}  & ECCVW'22  & 65.59 & 83.11 & 89.21 & -&-& 56.04 & 73.60 & 80.22 & - & -  & 46.70 & 70.00 & 78.80 & - & -\\
CFine\cite{b20}  & TIP'23  & 69.57 & 85.93 & 91.15 & - & - & 60.83 & 76.55 & 82.42 & - & - & 50.55 & 72.50 & 81.60 & - & -\\
PBSL\cite{b21}  & ACMMM'23 & 65.32 & 83.81 & 89.26 & - & - & 57.84 & 75.46 & 82.15 & - & - & 47.80 & 71.40 & 79.90 & - & -\\
BEAT\cite{b22}  & ACMMM'23  & 65.61 & 83.45 & 89.54 & - & - & 58.25 & 75.92 & 81.96 & - & - & 48.10 & 73.10 & 81.30 & - & -\\
LCR$^{2}$S\cite{b10}  & ACMMM'23 & 67.36 & 84.19 & 89.62 & 59.24 & - & 57.93 & 76.08 & 82.40 & 38.21 & - & 54.95 & 76.65 & 84.70 & 40.92 & -\\
IRRA\cite{b4}  & CVPR'23  & 73.38 & \underline{89.93} & 93.71 & 66.13 & 50.24 & 63.46 & \underline{80.25} & 85.82 & 38.06 & \underline{7.93} & 60.20 & \underline{81.30} & 88.20 & 47.17 & 25.28\\
RDE\cite{b6} & CVPR'24  & \textbf{75.94} & \textbf{90.14} & \textbf{94.12} & \underline{67.56} & \underline{51.44} & \textbf{67.68} & \textbf{82.47} & \textbf{87.36} & \underline{40.06} & 7.87& \textbf{65.35} & \textbf{83.95} & \textbf{89.90} & \textbf{50.88} & \textbf{28.08}\\
CSKT\cite{b34}  & ICASSP'24  & 69.70 & 86.92 & 91.80 & 62.74 & - & 58.90 & 77.31 & 83.56 & 33.87 & - & 57.75 & \underline{81.30} & \underline{88.35} & 46.43 & -\\
\hline
\textbf{Our AMNS} & -  & \underline{74.25} & 89.54 & \underline{93.83} & \textbf{67.67} & \textbf{53.16} & \underline{64.05} & 79.90 & \underline{85.90} & \textbf{41.27} & \textbf{10.36} & \underline{60.50} & \underline{81.30} & 87.65 & \underline{47.20} & \underline{25.38}\\
\Xhline{1px}
\end{tabular}
}
\label{tab.2}
\end{table*}

\section{Experimental section}
\subsection{Datasets and Settings}
\textbf{Datasets:} In our experiments, we evaluated the proposed methods on CUHK-PEDES\cite{b1}, ICFG-PEDES\cite{b38}, and RSTPReid\cite{b39} datasets. 

\textbf{Evaluation Protocols:} For all experiments, we use the popular Rank-K metric (K=1,5,10) to measure retrieval performance. We also use mean Average Precision (mAP) and mean Inverse Negative Penalty (mINP\cite{b15}) as auxiliary retrieval metrics to further evaluate the performance.

\textbf{Implementation Details:} In our experiments, we used the same version of  CLIP-ViTB/16 as IRRA\cite{b4}. Image augmentation includes color jitter, random horizontal flipping, random crop with padding, random grayscale, random erasing, and gaussian blur. Text augmentation involves random masking, replacement, and removal of word tokens. The model training parameters follow IRRA, the temperature parameter $\tau$ of BSDM is set to 0.02, $\gamma$ of WAF is set to 2, and the values of $\alpha$ and $\beta$ are set to 0.1 and 0.05, respectively. $\overline{\mathbf{E_i}}$ generates an attention mask with a momentum that starts from 0.996 and gradually increases to 1 with training, and the experiments show that it works best when the mask rate is 0.5. Specific experimental results are detailed in Section \ref{shiyan}.

\subsection{Experimental results}
\label{shiyan}
In Fig.~\ref{Fig.3}, the experimental results on AMNS show that the effect of attention-weighted selective mask based on attention is generally better than that of random mask, and the effect of attention-weighted selective mask reaches the optimal effect when the mask ratio = 0.5. Therefore, in the subsequent experiments, for the AWM strategy, we all set the mask ratio to 0.5 for the experiments.

As shown in Table~\ref{tab.1}, the SDM loss is the similarity distribution matching loss proposed by IRRA\cite{b4}, which is introduced in Section \ref{BSDM}. TAL loss is a novel triplet alignment loss proposed by RDE\cite{b6}, which effectively solves the problem of poor local minima and even model collapse under NC in the early stage of training and achieves better performance.The BSDM is the bidirectional similarity distribution matching loss proposed in this paper. Table~\ref{tab.1} shows that the use of BSDM loss in different methods can bring significant improvement in all aspects of performance, and the best performance is still achieved under the 20\% and 50\% noise rates, which fully indicates that the BSDM loss proposed by us is superior to the widely used TAL loss and SDM loss, and has better stability and robustness for NC.

Table~\ref{tab.2} shows that our method achieves good results in Rank metrics. The mAP index and mINP index achieve the best results in the first and second datasets. Higher mINP metrics mean that the correctly matched images are indexed at a more advanced position, which fully validates that our method is better at finding the correct matches and can be better used in real-life retrieval systems.
%自己方法的消融表格
\begin{table}[t]
\captionsetup{format=plain,justification=raggedright,singlelinecheck=false, font=small}
\caption{Ablation study on each component of AMNS on CUHK-PEDES.}
\centering
\setlength{\tabcolsep}{3pt} % 调整列间距
\fontsize{8}{9}\selectfont % 设置字体大小为8pt，行距为9pt
\resizebox{\columnwidth}{!}{ % 使用 \columnwidth 以填充整列
\begin{tabular}{c|c|ccc|ccccc}
    \Xhline{1px}
    & & \multicolumn{3}{c|}{Components} & \multicolumn{5}{c}{CUHK-PEDES} \\
    No.& Methods & BSDM & WAF & L$_{id}$ & Rank-1 & Rank-5 & Rank-10 & mAP & mINP \\
    \hline
    0 & \multicolumn{1}{m{1.7cm}|}{Baseline} & & & & 69.12 & 87.02 & 92.02 & 62.17 & 46.08 \\
    1 & \multicolumn{1}{m{1.7cm}|}{+BSDM} & \checkmark & & &73.34 & 89.23 & 93.02 & 65.25 & 48.76 \\
    2 & \multicolumn{1}{m{1.7cm}|}{+WAF}  & &\checkmark & & 71.74 & 88.09 & 92.56 & 64.94 & 49.35 \\
    3 & \multicolumn{1}{m{1.7cm}|}{+L$_{id}$}& & & \checkmark & 66.57 & 85.62 & 91.00 & 61.12 & 45.82 \\
    4 &\multicolumn{1}{m{1.7cm}|}{+WAF+L$_{id}$}&  & \checkmark & \checkmark & 69.83 & 86.11 & 91.55 & 65.28 & 52.07 \\
    5 &\multicolumn{1}{m{1.7cm}|}{+BSDM+WAF}& \checkmark & \checkmark &  & 73.60 & 88.81 & 93.11 & 65.56 & 49.21 \\
    6 & \multicolumn{1}{m{1.7cm}|}{+BSDM+L$_{id}$}& \checkmark & & \checkmark & 74.16 & 89.02 & 93.52 & 67.37 & 52.57  \\
    \hline
    7 &\multicolumn{1}{m{1.5cm}|}{AMNS}& \checkmark&\checkmark & \checkmark & \textbf{74.25} & \textbf{89.54} & \textbf{93.83} & \textbf{67.67} & \textbf{53.16} \\
    \Xhline{1px}
\end{tabular}
}
\label{tab.3}
\end{table}
\subsection{Ablation experiments}
In this section, we fine-tune the CLIP-ViT-B/16 model with AWM loss as a baseline and perform ablation studies on the CUHK-PEDES dataset.

\textbf{Effectiveness of BSDM:} As can be seen from No.1 in Table~\ref{tab.3}, just by adding BSDM to Baseline, the accuracies of Rank-1, Rank-5, and Rank-10 are improved by 4.22\%, 2.21\%, and 1.00\%, respectively, and the values of mAP and mINP are improved by 3.08\% and 2.68\%, respectively, which reflect the effectiveness of BSDM in the Person Retrieval task.

\textbf{Complementarity between BSDM and WAF:} As can be seen from No.5 in Table~\ref{tab.3}, the addition of WAF improves the Rank1 metric by 0.26\%, and the mAP and mINP metrics by 0.31\% and 0.45\%, respectively, which fully demonstrates that WAF further improves the model's ability to recognize hard samples when it is complemented with BSDM.

\section{Conclusion}
In this paper, we explore Noise Correspondence (NC) and the noisy matching problem. We introduce the Attention-Weighted Selective Mask (AWM) to alleviate the noise matching problem and propose a novel noise suppression method, Bidirectional Similarity Distribution Matching (BSDM), combined with an improved focal loss, to address the Noisy Correspondence (NC) issue. Together, these methods achieve excellent performance. Extensive experiments have been conducted, confirming the superiority of our approach in resolving the aforementioned issues and demonstrating its ability to find the correct matches better.

\end{document}